\pdfoutput=1

\documentclass[11pt]{article}

\usepackage[final]{acl}

\usepackage{soul}
\usepackage{booktabs}
\usepackage{times}
\usepackage{latexsym}
\usepackage{amsmath}
\usepackage{multirow}
\DeclareMathOperator*{\argmax}{arg\,max}

\usepackage[T1]{fontenc}

\usepackage[utf8]{inputenc}

\usepackage{microtype}

\usepackage{inconsolata}

\usepackage{graphicx}
\usepackage{enumitem} 
\usepackage{soul}
\renewcommand{\hl}[1]{}
\newif\ifshowcontent
\showcontenttrue

\usepackage{ifthen}

\newcommand{\docmode}{submit} 

\newcommand{\showcontent}[2]{%
  \ifthenelse{\equal{\docmode}{#1}}{#2}{}%
}
\title{REIC: RAG-Enhanced Intent Classification at Scale}
\author{Ziji Zhang, Michael Yang, Zhiyu Chen, Yingying Zhuang, Shu-Ting Pi, \\
\textbf{Qun Liu, Rajashekar Maragoud, Vy Nguyen, Anurag Beniwal} \\
Amazon.com Inc, Seattle, USA \\
\texttt{\{czhangzi,abyang,zhiyuche,yyzhuang,shutingp\}@amazon.com}\\ 
\texttt{\{qunliu,maragoud,nguynvy,beanurag\}@amazon.com}\\ 
}

\begin{document}
\maketitle
\begin{abstract}
Accurate intent classification is critical for efficient routing in customer service, ensuring customers are connected with the most suitable agents while reducing handling times and operational costs. However, as companies expand their product lines, intent classification faces scalability challenges due to the increasing number of intents and variations in taxonomy across different verticals. 
In this paper, we introduce \textbf{REIC}, a \textbf{R}etrieval-augmented generation \textbf{E}nhanced \textbf{I}ntent \textbf{C}lassification approach, which addresses these challenges effectively. 
REIC leverages retrieval-augmented generation (RAG) to dynamically incorporate relevant knowledge, enabling precise classification without the need for frequent retraining. Through extensive experiments on \showcontent{review}{synthetic datasets}\showcontent{submit}{real-world datasets}, we demonstrate that REIC outperforms traditional fine-tuning, zero-shot, and few-shot methods in large-scale customer service settings. Our results highlight its effectiveness in both in-domain and out-of-domain scenarios, demonstrating its potential for real-world deployment in adaptive and large-scale intent classification systems.
\showcontent{review}{\\\textbf{Disclaimer: We use only synthetic datasets in this paper, and the numbers reported do not correspond to any company products.}}
\end{abstract}


\section{Introduction}

Customer service~\cite{cui2017superagent,chen-etal-2024-identifying, nguyen-etal-2020-dynamic,qi-etal-2021-benchmarking,chen-etal-2023-generate,zhou-etal-2023-towards-open,pi2024topology} is critical for modern e-commerce but also one of the most resource-intensive departments. Different agents, either human or model, are trained to handle specific types of customer issues, making precise intent classification, particularly at the issue level, crucial for efficient routing. High issue-oriented intent accuracy ensures that customers are connected with the most suitable agents, reducing unnecessary transfers and lowering handling times. This optimization not only enhances customer satisfaction but also cuts operational costs by streamlining interactions and improving overall service efficiency. For model-based automatic resolvers in chatbot agentic systems \cite{gupta2024dard}, the ability to precisely identify user intent is essential for delivering contextually appropriate and solution-oriented responses.

\begin{figure}[th]
    \centering
   \includegraphics[width= 1.00\linewidth]{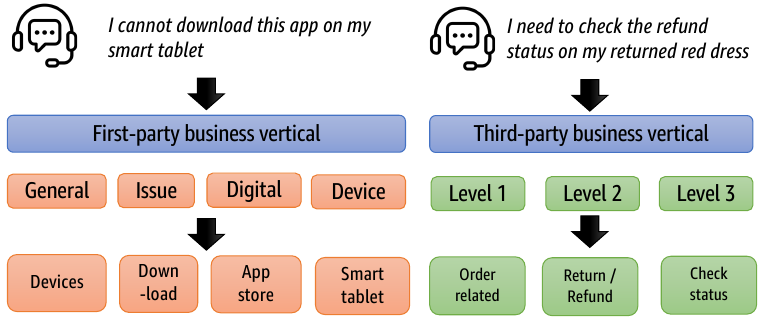}
    \caption{We present the heterogeneous intent structure with representative examples, illustrating the intent label hierarchy in each vertical.}
    \label{fig:label}
\end{figure}

As companies expand their product lines, intent classification faces two key challenges. First, the number of customer intents grows over time, requiring models to adapt to new intents quickly. Second, intent taxonomies can vary across product lines, making it difficult to maintain a unified classification system\cite{pi2024universal}. For example we organize products into different verticals in e-commerce: with third-party products, customer usually inquire about physical retail orders or consumer accounts, and intents are categorized into three levels from coarse to fine-grained. In contrast, first-party products require more customized customer services due to our proprietary device and digital product offerings. For instance, a customer seeking device troubleshooting may interact with an agent who can access real-time diagnostic information and perform specific troubleshooting steps on the user's behalf. This necessitates a broader set of intent categories to accommodate diverse customer needs, as illustrated in Figure~\ref{fig:label}. This heterogeneity complicates intent classification, demanding scalable and flexible approaches to ensure accurate routing and efficient customer service. In this work we demonstrate our method using two verticals but it can be easily adapted to more.

In this paper, we propose a novel \textbf{R}etrieval-augmented generation \textbf{E}nhanced \textbf{I}ntent \textbf{C}lassification (\textbf{REIC}) approach that reduces computational complexity and improves scalability for intent classification.  We demonstrate the effectiveness of this approach through extensive experiments on
\showcontent{review}{\textbf{synthetic datasets}}\showcontent{submit}{real-world datasets}
, showing that our method outperforms traditional fine tuning and zero-shot or few-shot methods in large-scale customer service settings. Our results on both in-domain and out-of-domain intents demonstrate its potential to improve classification accuracy and enable dynamic updates without retraining, making it ideal for industry-scale applications.

\section{Related Work}
\subsection{Intent classification}
Early work on intent classification for dialogues often relied on bag-of-words or recurrent models. For example, \citet{schuurmans2019intent} evaluated various classifiers on a multi-domain intent dataset and found that a simple SVM with hierarchical label taxonomy outperformed deeper LSTM models. With the advance of transformer architectures, researchers began to leverage self-attention and multi-task learning for intent understanding. \citet{ahmadvand2020jointmap} introduced a joint intent mapping model that simultaneously classifies high-level intent and maps queries to fine-grained product categories. \citet{wang2021mell} employed a slowly updated text encoder and global/local memory networks to mitigate catastrophic forgetting and parameter explosion for large-scale intent detection task. Recent work has pushed toward using large pre-trained models and retrieval-based prompting to enable cross-domain and zero/few-shot intent classification. \citet{liu2024lara} proposed a framework which integrates a fine-tuned XLM-based intent classifier with an LLM to essentially treat multi-turn intent understanding as a zero-shot task. \citet{yu2021few} also explored retrieval-based
methods for intent classification and slot filling tasks in few-shot settings. Our work adopts similar in-context learning (ICL) setup while focusing on handling large-scale multi-domain intent classification task from industry level applications. 

\subsection{In-context Learning}
The performance of LLM has been significantly enhanced in few-shot and zero-shot NLP tasks through ICL. Recent ICL research focus on how to effectively identify and interpret retrieved context. \citet{guu2020realmretrievalaugmentedlanguagemodel} first showed how to pre-train masked language models with a knowledge retriever in an unsupervised manner. \citet{karpukhin2020densepassageretrievalopendomain} proposed a training pipeline in which retrieval is implemented using dense representations alone and embeddings are learned from a small number of questions and passages with a dual-encoder. \citet{ram2023incontextretrievalaugmentedlanguagemodels} considered simple alternatives to only prepend retrieved grounding documents to the input, instead of modifying the LLM architecture to incorporate external information. Similar approaches have proven particularly effective in the application of RAG on dialogue systems \cite{shuster2022blenderbot3deployedconversational,shuster-etal-2022-language}, specifically goal-oriented and domain-specific dialogs from customer service scenarios \cite{zhuang-etal-2021-weakly, 10.1145/3543873.3587680}. In our work, we utilize ICL in both LLM fine-tuning stage for data generation and at inference-time with an intent candidate retriever.
\begin{figure*}[th]
    \centering
   \includegraphics[width= 1.00\linewidth]{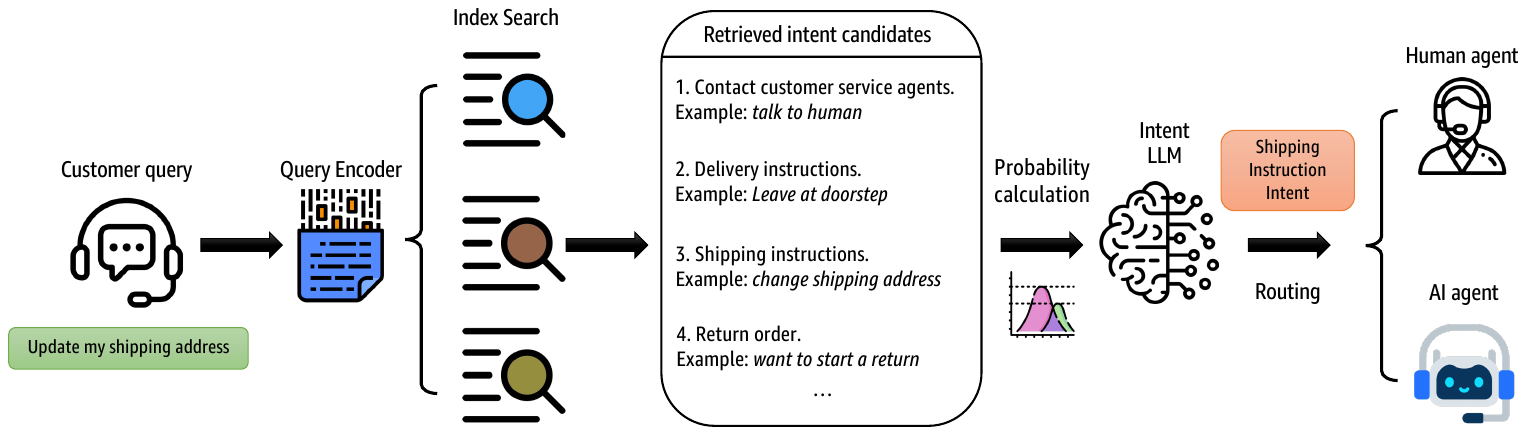}
    \caption{The proposed REIC method from customer query to routing intent leveraging on vector retrieval and probability calculation.}
    \label{fig:rag}
\end{figure*}

\section{Preliminary}
Intent classification for queries is typically framed as a multiclass text classification problem. Specifically, given a customer query $q\in Q$, the goal is to map it to one of the $k$ pre-defined intents $t\in T=\{t_1,...,t_k\}$ using a model $\mathcal{M}$ so that the predicted intent $\hat{t} = \mathcal{M}(q)$ maximizes the probability of correctly classifying $q$. Formally, this can be expressed as:
\begin{equation}
    \hat{t} = \arg\max_{t \in T} \, P(t \mid q; \theta)
\end{equation}
where $P(t \mid q; \theta)$ denotes the probability of intent $t$ given query $q$, parameterized by $\theta$ of the model $\mathcal{M}$.

However, intent labels are often not mutually independent; correlations between intents can exist, making a flat classification structure suboptimal. This leads to two major challenges in large-scale, industry-level intent classification: 1) \textbf{scalability}: a large number of intent labels $k$ can make flat classification computationally expensive and difficult to scale, especially as the set of intents grows; 2) \textbf{label correlation}:
related intents, such as "Order Issue" and its sub-intents ``Track Order'' or ``Cancel Order'', are treated independently in flat classification, ignoring their hierarchical relationships and increasing the risk of misclassification.

To address these issues, hierarchical intent classification is preferred. In this approach, a query $q$ is classified progressively from general categories to specific sub-intents, enhancing both efficiency and accuracy for industry-scale applications.

Note that for a more generalized setting, intents from different verticals and domains may have entirely different hierarchy and ontology. In Figure \ref{fig:label}, we demonstrated some examples of intent hierarchy in our application, which involves customer service query intent detection with two business verticals: \textbf{Third-Party} or \textbf{3P} business (customer contacting about third-party physical retail orders or consumer accounts) and \textbf{First-Party} or \textbf{1P} business (customer contacting about first-party digital or device issues). 
Both verticals span a diverse range of product types, reflecting the broad scope of customer inquiries handled by our system. 
If we use the traditional single-head flattened intent labels, the total intent ontology set size would be at $10^3$ level which create major challenges for accurate intent classification. By creating hierarchical intent ontology across different business verticals, each classification head only needs to handle less than 50 intents that are more manageable for language models. In the following sections of this paper, we utilize this intent ontology setup for experiments and comparisons.
\section{Method}\label{sec:method}
LLMs has revolutionized the landscape of customer engagement, particularly in the domain of intent detection systems. While these models demonstrate remarkable capabilities in language comprehension and knowledge representation, their adaptation to domain-specific contexts presents notable challenges. Specifically, the integration of industry-specific terminology, organizational nomenclature, and distinctive customer service scenarios necessitates fine-tuning and customization of these models. To address these limitations and enhance the accuracy of customer intent identification, we introduce a novel approach REIC for RAG-Enhanced Intent Classification. 

REIC aims to bridge the gap between the generalized capabilities of LLMs and the specialized requirements of diverse business environments. By leveraging RAG, our approach seeks to enhance the precision and relevance of intent detection, thereby facilitating more nuanced and context-aware customer interactions across various industry-specific scenarios. The method consists of three main components: index construction, candidate retrieval, and intent probability calculation (Figure \ref{fig:rag}).

\paragraph{Index Construction} We first construct a dense vector index containing (query, intent) pairs from a held-out annotated dataset. Each query is encoded using a pre-trained sentence transformer model to generate dense vector representations. The corresponding intent labels are stored alongside these embeddings. The intent labels follow a hierarchical structure with $d$ dimensions which represents different intent domain knowledge and might range from different domains.

\paragraph{Candidate Retrieval} Given a new query $q$, we first encode it using the same encoder for index construction to obtain its dense vector representation $\mathbf{v}_q$. We then perform approximate nearest neighbor search to retrieve the top-k most similar (query, intent) pairs, denoted as set $E$. The similarity is computed using cosine distance between the query vector and indexed vectors:
\begin{equation}
    sim(q, q_i) = \frac{\mathbf{v}_q \cdot \mathbf{v}_i}{\|\mathbf{v}_q\| \|\mathbf{v}_i\|}
\end{equation}
where $\mathbf{v}_i$ represents the vector encoding of the i-th indexed query.
\paragraph{Intent Probability} For the retrieved set $E$, we leverage a fine-tuned LLM $\mathcal{M}$ to perform constrained decoding and calculate the probabilities of the possible intents. Given a prompt template $\mathcal{P}$, the LLM takes as input the instantiated prompt, which includes the original query $q$ and the retrieved (query, intent) pairs as context. For each unique intent $t_j$ in $E$, we compute:
\begin{equation}
    P(t_j|q, E) = \mathcal{M}(\mathcal{P}, q, E)_{t_j}
\end{equation}
where $\mathcal{M}(\mathcal{P}, q, E)_{t_j}$ represents the model's predicted probability for $t_j$ given the query and retrieved examples.\\
The final intent classification is determined by selecting the intent with the highest probability:
\begin{equation}
    \hat{t} = \argmax_{t_j \in E} P(t_j|q, E)
\end{equation}
This approach enables dynamic updates to the intent space by simply adding new (query, intent) pairs to the index, leveraging the in-context learning capabilities of the LLM without requiring model retraining.
\begin{figure}[th]
    \centering
   \includegraphics[width= 1.00\linewidth]{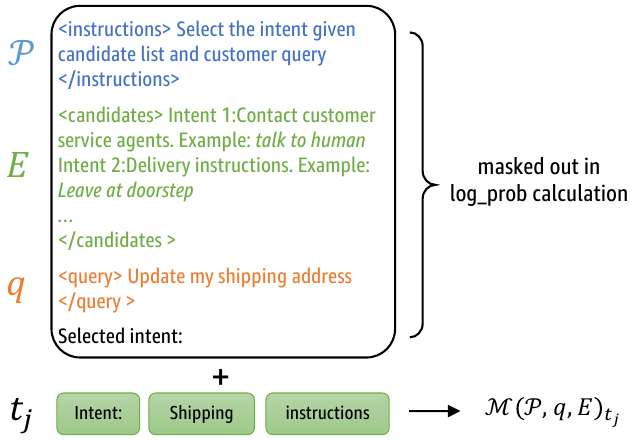}
    \caption{Constrained decoding for probability calculation.}
    \label{fig:probs}
\end{figure}

\hl{yeah we can merge it with intent probability calculation/generation} The probability-based reranking helps mitigate potential LLM hallucination by grounding predictions in retrieved examples (Figure \ref{fig:probs}).\hl{why not direct generation} With tradition greedy decoding, sometimes the LLM might generate intents outside of the given candidate list and cause downstream routing failure. We perform constrained decoding to calculate the probability of each retrieved intent $t_j$ in $E$, which ensures the success of downstream routing. Given prompt $\mathcal{P}$ with instructions, retrieved candidates $E$, and customer query $q$, we append $t_j$ at the end to calculate the total logits from model forward pass $\mathcal{L}_{t_j} = \mathcal{M}(\mathcal{P}(E, q) + t_j)$. Then we mask out the positions of $\mathcal{P}(E, q)$ and accumulate the log probabilities for the intent sequence $t_j$ with length $s_j$:
\begin{equation}
    \mathcal{M}(\mathcal{P}, q, E)_{t_j} = \exp{(\sum_{t_j}\text{LogSoftmax}(\mathcal{L}_{t_j})/s_j)}
\end{equation}
During training, we train the intent LLM $\mathcal{M}$ by minimizing the cross-entropy loss between the predicted and ground-truth intents. During inference, instead of traditional auto-regressive next token decoding, we perform one model forward-pass calculation with a batch size $k$ for top-k intent candidates and get the $k$ probabilities for re-ranking and final intent prediction.
\section{Experimental Setup}

\subsection{Datasets}\label{sec:data}

Due to business considerations, we are not permitted to share the results using the original customer data. As a result, we manually anonymized both the labels and transcripts to ensure no personal information is included. Additionally, specific product and service names were denonymized to prevent the identification of the company from the transcript or label descriptions. Despite these modifications, the conclusions drawn from our experiments remain valid. 
The final dataset contains 52,499 training samples with 35,041 \textbf{1P Business} queries and 17,458 \textbf{3P Business} queries. The test set consists of 3,647 \textbf{1P Business} queries and 1,717 \textbf{3P Business} queries respectively using random sampling. All of the data samples have incorporated retrieved intent candidates from the retriever. We also performed dataset cleaning in the training set to make sure the true intent is contained in the retrieved list. During inference, we use the actual noisy retrieved list which also relies on the capability of the embedding model. 

\begin{table*}[!ht]
\small
   \centering
   \begin{tabular}{c|c|c|c|c|c|c|c|c|c}
   \toprule
   \multirow{2}{*}{Models} & \multicolumn{3}{c|}{3P Business vertical} & \multicolumn{3}{c|}{1P Business vertical } & \multicolumn{3}{c}{Overall} \\
      & Precision & Recall & F1 & Precision & Recall & F1 & Precision & Recall & F1 \\
   \midrule
   RoBERTa & 0.527 & 0.447 & 0.483 & 0.583 & 0.488 & 0.531 & 0.565 & 0.474 & 0.516 \\
   Mistral Classification & 0.215 & 0.228 & 0.221 & 0.301 & 0.250 & 0.273 & 0.269 & 0.243 & 0.255\\
   Claude Zero-shot & 0.338 & 0.250 & 0.287 & 0.238 & 0.170 & 0.199 & 0.271 & 0.196 & 0.227\\
   Claude Few-shot & 0.386 & 0.289 & 0.331 & 0.350 & 0.308 & 0.328 & 0.361 & 0.302 & 0.329\\
   Claude + RAG & 0.473 & 0.438 & 0.455 & 0.415 & 0.389 & 0.402 & 0.434 & 0.404 & 0.419\\
   \midrule
   \textbf{REIC} & \textbf{0.538} & \textbf{0.546} & \textbf{0.542} & \textbf{0.600} & \textbf{0.574} & \textbf{0.587} & \textbf{0.579} & \textbf{0.565} & \textbf{0.572}\\
   \bottomrule
   \end{tabular}\caption{
   Intent detection confusion matrix on different business with different methods}
   \label{t:main}
   \end{table*}
   
\subsection{Compared Methods}
\label{sec:baselines}
We consider the following baselines: 
\begin{itemize}[leftmargin=*]
\setlength{\itemsep}{0pt}
\setlength{\parskip}{0pt}
\setlength{\parsep}{0pt}
\setlength{\topsep}{0pt} 
\item \textbf{RoBERTa}: We fine-tune RoBERTa-base~\cite{liu2019roberta} with multiple classification heads. This adaptation allowed the model to simultaneously categorize utterances across multiple dimensions.
\item \textbf{Mistral Classification}\footnote{Due to legal concerns, we are not permitted to use non-commercial LLMs like Llama.}: We fine-tune a Mistral-7B-v0.3\footnote{https://huggingface.co/mistralai/Mistral-7B-v0.3} with a sequence classification head. Instead of directly generating output sequences, the model projects the pooled embedding into a space with the same dimension as the number of classes. 
\item \textbf{Claude Zero-shot}: We employ the Claude 3.5 Sonnet model in a zero-shot configuration. To facilitate accurate intent prediction, we craft a comprehensive prompt that explicitly defines each potential intent. 
\item \textbf{Claude Few-shot}: Similar to \textbf{Claude Zero-shot}, we incorporate 20 demonstration examples, with 10 from each vertical, to enhance coverage of diverse intents across different domains.
\item \textbf{Claude+RAG}: Instead of using a fine-tuned LLM, we employ Claude 3.5 Sonnet as the backbone and incorporate the same set of retrieved candidates as described in Section~\ref{sec:method} into the prompt. This comparison allows us to assess whether a smaller fine-tuned LLM can perform competitively against a large foundation model for this task.
\end{itemize}



%

\subsection{Implementation Details}\label{sec:details}
The LLM component of our REIC approach utilizes a fine-tuned model from Mistral-7B-Instruct-v0.2\footnote{https://huggingface.co/mistralai/Mistral-7B-Instruct-v0.2}.  
We applied 8 NVIDIA-A100 40GB GPUs with 96 vCPUs to conduct PEFT \cite{peft} training with LoRA adapters \cite{hu2022lora}. We choose a set of LoRA parameters with a rank of 8, an alpha value of 16, and a dropout rate of 0.1. The training batch size is set to 8 per GPU with a learning rate of $2e^{-5}$. We train the model using Cross Entropy Loss for 3 epochs which takes around 3 hours on the instance.

We experimented with the following off-the-shell retrievers for candidate retrieval:
\begin{itemize}[leftmargin=*]
\setlength{\itemsep}{0pt}
\setlength{\parskip}{0pt}
\setlength{\parsep}{0pt}
\setlength{\topsep}{0pt}  
    \item \textbf{BM25}~\cite{robertson1995okapi} is a widely used traditional sparse retrieval method. Although it is unsupervised, it consistently demonstrates strong performance across a variety of benchmarks\cite{thakur2021beir}.
    
    \item \textbf{MPNet}\footnote{https://huggingface.co/sentence-transformers/all-mpnet-base-v2}~\cite{song2020mpnet} is a sentence embedding model fine-tuned on one billion sentence pairs using a contrastive learning objective.
    \item \textbf{Contriever-MS MACRO}~\cite{izacard2022unsupervised} is an unsupervised dense retriever pre-trained with contrastive learning and fine-tuned on MS MARCO~\cite{nguyen2016ms}.
    \item \textbf{ColBERT-v2}~\cite{santhanam-etal-2022-colbertv2} is a late-interaction retriever that combines denoised supervision and residual compression to improve retrieval quality and reduce space footprint.
\end{itemize}


\section{Results} 





\subsection{Intent Detection Ability}

To evaluate the effectiveness of our REIC method in intent detection, we conducted experiments comparing it against several baseline methods described in \S\ref{sec:baselines}. Our results, presented in Table \ref{t:main}, illustrate performance across two business verticals (\textit{3P Business} and \textit{1P Business}) and an overall aggregate assessment based on Precision, Recall, and F1-score.

The results indicate that our REIC offers significant advantages in intent detection over standard fine-tuning or prompting-based methods. While fine-tuned models like RoBERTa perform reasonably well, they require extensive retraining when new intents emerge. We hypothesize that the limited performance of the Mistral Classification model stems from its nature as a decoder-only architecture, which may be less effective in extracting the semantic meaning of input query. Additionally, since it is not pretrained for classification tasks, incorporating a classification head during fine-tuning is unlikely to yield optimal results. Prompting-based approaches (Claude Zero-shot and Few-shot) generally underperformed, with Claude Few-shot achieving a maximum F1-score of 0.329 overall. The Claude + RAG method improved performance compared to standalone prompting but remained inferior to our approach by 26.7\%. 

These observations confirm that the integration of RAG and fine-tuned LLM enables greater flexibility, improved precision, and higher recall rates, making it well-suited for handling diverse and evolving intent spaces in different applications. 

\subsection{Impact of Retrievers}

In order to evaluate the impact of retrievers on the final performance, we experimented four different retrievers in REIC including one sparse retrieval method and three dense retrievers, details in \S\ref{sec:details}. The intent detection accuracy across different business verticals using these retrievers is presented in Table~\ref{table2}.

BM25, despite being an unsupervised sparse retrieval method, performs competitively, achieving an overall accuracy of 0.532. Among the dense retrievers, MPNet outperforms the others, attaining the highest accuracy across both the 3P Business and 1P Business verticals. This suggests that MPNet’s contrastive learning-based sentence embeddings are highly effective for retrieving relevant candidates that aid intent classification. 
In contrast, Contriever exhibits the lowest accuracy across all categories.

Our findings show that retriever selection significantly impacts intent classification. Although BM25 is a strong baseline, dense retrievers like MPNet consistently outperform it. This highlights the value of high-quality embeddings and extensive fine-tuning on large datasets, which is why we have chosen MPNet as our final retriever in REIC. 
\begin{table}[!t]
   \centering
   \resizebox{0.85\columnwidth}{!}{
   \begin{tabular}{@{}cccc@{}}
\toprule
Retriever  & 3P Business    & 1P Business    & Overall        \\ \midrule
BM25       & 0.521          & 0.537          & 0.532          \\
Contriever & 0.450          & 0.461          & 0.457          \\
ColBERTv2  & 0.503          & 0.560          & 0.542          \\
MPNet      & \textbf{0.545} & \textbf{0.573} & \textbf{0.564} \\ \bottomrule
\end{tabular}}
   \caption{
   Intent detection accuracy on different business verticals using different retrievers in REIC}\label{table2}
   \end{table}

\section{Impact of Retrieval Candidate Size} \label{appendix:topk}
\begin{figure}[th]
    \centering
   \includegraphics[width= 0.950\linewidth]{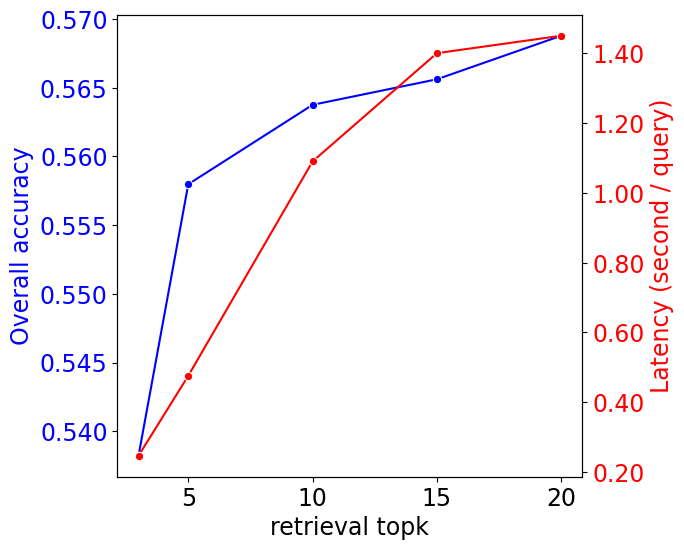}
    \caption{The accuracy and latency when using different retrieval top-k values.}
    \label{fig:topk}
\end{figure}

We investigated the impact of different retrieval candidate numbers (top-k) in REIC to balance intent detection accuracy and inference latency. The Figure \ref{fig:topk} illustrates the trade-off between these two factors, with overall accuracy plotted on the left y-axis (blue) and inference latency on the right y-axis (red) against different values of top-k. From the accuracy perspective, increasing top-k allows the model to access a broader range of relevant information, leading to better predictions. Beyond a certain threshold, additional retrieved candidates contribute minimally to accuracy while still increasing computational complexity. Latency, on the other hand, exhibits a sharp rise as top-k increases. This indicates a crucial trade-off: although retrieving more candidates can improve accuracy, it also leads to longer inference times, which may not be suitable for real-time applications.

In our experiments, we select top-k = 10 which ensures a meaningful accuracy boost without incurring excessive computational costs. However, the ideal top-k may vary depending on application requirements. For instance, real-time systems such as customer service chatbots or voice assistants may favor a lower top-k to maintain fast response times. Conversely, offline or batch-processing applications could accommodate higher top-k values if maximizing accuracy is a priority. Our findings emphasize the need to carefully tune retrieval parameters in REIC to meet specific operational demands.

\subsection{Robustness on Unseen Intents}
To evaluate REIC's robustness on unseen intents, we trained our models exclusively on the \textit{3P Business} vertical and tested them on the \textit{1P Business} vertical, simulating a real-world out-of-domain scenario. As illustrated in Figure \ref{fig:label}, \textit{1P Business} vertical has 4 intent category with more than 800 unique intent combinations, while the training data used from \textit{3P Business} vertical has 3 intent category with only around 70 unique intents. This out-of-domain scenario helps assess how well REIC generalizes to new, previously unseen intents. The results are summarized in Table~\ref{t:out-domain}.

Claude Zero-shot performs the worst, with an accuracy of 0.17 on the 1P Business vertical. In contrast, Claude Few-shot shows improvement, achieving an accuracy of 0.308 on the 1P Business vertical. This demonstrates that providing a few examples significantly enhances the model’s ability to generalize. 

Notably, the RAG-based methods, particularly Claude + RAG, significantly outperform Claude Few-shot, achieving 0.389 on the 1P Business vertical. This demonstrates the advantage of our RAG-based strategy in handling unseen intents, as it dynamically retrieves the most relevant examples to enhance predictions, surpassing static few-shot examples. Similarly, REIC, although slightly lower than Claude + RAG, still performs strongly compared to Claude Few-shot, highlighting the model’s effectiveness on both in-domain and out-of-domain intents. Overall, REIC excels in-domain, and its performance on the 1P Business vertical remains competitive with Claude Few-shot, underscoring the robustness and adaptability of our REIC approach for unseen domains.

\begin{table}[ht]
    \centering
    \resizebox{0.9\columnwidth}{!}{
    \begin{tabular}{@{}cccc@{}}
    \toprule
    Models           & 3P Business    & 1P Business        & Overall        \\ \midrule
    Claude Zero-shot & 0.250          & 0.170          & 0.196          \\
    Claude Few-shot  & 0.289          & 0.308          & 0.302          \\
    Claude + RAG     & 0.438          & \textbf{0.389} & \textbf{0.404} \\
    REIC             & \textbf{0.538} & 0.283          & 0.364          \\ \bottomrule
    \end{tabular}}    \caption{
    Out-of-domain intent detection accuracy}
    \label{t:out-domain}
\end{table}

\showcontent{submit}{\subsection{Online Deployment Performance}
Following the deployment of our REIC in an internal system, we observed significant improvements in online performance, particularly in intent detection accuracy. Compared to the previously deployed system, which relies on two separate models using traditional fine-tuning approaches for different business verticals, our REIC method simplified the system using just one consolidated model, which reduced misclassifications and improved resolution routing. To measure the success rate, we include a confirmation question for the customer to verify whether the predicted intent or issue is correct. Our REIC leads to a 3.38\% absolute improvement for customer positive response rate. These improvements validate the efficacy of REIC for real-world deployment, offering both accuracy gains and operational efficiency in intent classification.}

\hl{limitation: calibration of probability score; retrieval improvements, can try fine-tune; intent label confusion overlap}

\section{Conclusion}

This paper presents a novel RAG-Enhanced Intent Classification (REIC) method that addresses scalability challenges and the heterogeneity of intent taxonomies in large-scale customer service systems. By incorporating a hierarchical intent classification strategy, REIC significantly reduces computational complexity. Leveraging the RAG technique, our method dynamically integrates contextually relevant retrieved examples, outperforming traditional fine-tuning, as well as zero-shot and few-shot approaches, in intent detection tasks. Additionally, our results demonstrate strong performance on both in-domain and out-of-domain test sets, highlighting its applicability for industry-scale applications.

\section{Limitations}

While REIC demonstrates strong performance in both in-domain and out-of-domain intent classification, it has a few limitations. Its accuracy remains closely tied to the quality of the retriever: if the correct intent is not among the retrieved candidates, the model cannot recover, underscoring the need for more robust retrieval methods or fallback mechanisms. In addition, although REIC allows for dynamic updates without retraining, it still relies on a fixed number of retrieved candidates, creating a trade-off between accuracy and latency that may hinder its deployment in real-time applications. Future work could address these challenges by developing adaptive retrieval strategies or introducing confidence-based mechanisms to dynamically adjust the candidate pool.
\clearpage



\bibliography{custom}

@inproceedings{chen-etal-2023-generate,
    title = "Generate-then-Retrieve: Intent-Aware {FAQ} Retrieval in Product Search",
    author = "Chen, Zhiyu  and
      Choi, Jason  and
      Fetahu, Besnik  and
      Rokhlenko, Oleg  and
      Malmasi, Shervin",
    editor = "Sitaram, Sunayana  and
      Beigman Klebanov, Beata  and
      Williams, Jason D",
    booktitle = "Proceedings of the 61st Annual Meeting of the Association for Computational Linguistics (Volume 5: Industry Track)",
    month = jul,
    year = "2023",
    address = "Toronto, Canada",
    publisher = "Association for Computational Linguistics",
    url = "https://aclanthology.org/2023.acl-industry.73/",
    doi = "10.18653/v1/2023.acl-industry.73",
    pages = "763--771",
}

@inproceedings{chen-etal-2024-identifying,
    title = "Identifying High Consideration {E}-Commerce Search Queries",
    author = "Chen, Zhiyu  and
      Choi, Jason Ingyu  and
      Fetahu, Besnik  and
      Malmasi, Shervin",
    editor = "Dernoncourt, Franck  and
      Preo{\c{t}}iuc-Pietro, Daniel  and
      Shimorina, Anastasia",
    booktitle = "Proceedings of the 2024 Conference on Empirical Methods in Natural Language Processing: Industry Track",
    month = nov,
    year = "2024",
    address = "Miami, Florida, US",
    publisher = "Association for Computational Linguistics",
    url = "https://aclanthology.org/2024.emnlp-industry.42/",
    doi = "10.18653/v1/2024.emnlp-industry.42",
    pages = "563--572",
}

@inproceedings{
    thakur2021beir,
    title={{BEIR}: A Heterogeneous Benchmark for Zero-shot Evaluation of Information Retrieval Models},
    author={Nandan Thakur and Nils Reimers and Andreas R{\"u}ckl{\'e} and Abhishek Srivastava and Iryna Gurevych},
    booktitle={Thirty-fifth Conference on Neural Information Processing Systems Datasets and Benchmarks Track (Round 2)},
    year={2021},
    url={https://openreview.net/forum?id=wCu6T5xFjeJ}
}

@article{robertson1995okapi,
  title={Okapi at TREC-3},
  author={Robertson, Stephen E and Walker, Steve and Jones, Susan and Hancock-Beaulieu, Micheline M and Gatford, Mike and others},
  journal={Nist Special Publication Sp},
  volume={109},
  pages={109},
  year={1995},
  publisher={National Instiute of Standards \& Technology}
}

@inproceedings{santhanam-etal-2022-colbertv2,
    title = "{C}ol{BERT}v2: Effective and Efficient Retrieval via Lightweight Late Interaction",
    author = "Santhanam, Keshav  and
      Khattab, Omar  and
      Saad-Falcon, Jon  and
      Potts, Christopher  and
      Zaharia, Matei",
    editor = "Carpuat, Marine  and
      de Marneffe, Marie-Catherine  and
      Meza Ruiz, Ivan Vladimir",
    booktitle = "Proceedings of the 2022 Conference of the North American Chapter of the Association for Computational Linguistics: Human Language Technologies",
    month = jul,
    year = "2022",
    address = "Seattle, United States",
    publisher = "Association for Computational Linguistics",
    url = "https://aclanthology.org/2022.naacl-main.272/",
    doi = "10.18653/v1/2022.naacl-main.272",
    pages = "3715--3734"
}

@article{nguyen2016ms,
  title={Ms marco: A human-generated machine reading comprehension dataset},
  author={Nguyen, Tri and Rosenberg, Mir and Song, Xia and Gao, Jianfeng and Tiwary, Saurabh and Majumder, Rangan and Deng, Li},
  year={2016}
}

@article{
izacard2022unsupervised,
title={Unsupervised Dense Information Retrieval with Contrastive Learning},
author={Gautier Izacard and Mathilde Caron and Lucas Hosseini and Sebastian Riedel and Piotr Bojanowski and Armand Joulin and Edouard Grave},
journal={Transactions on Machine Learning Research},
issn={2835-8856},
year={2022},
url={https://openreview.net/forum?id=jKN1pXi7b0},
note={}
}

@article{song2020mpnet,
  title={Mpnet: Masked and permuted pre-training for language understanding},
  author={Song, Kaitao and Tan, Xu and Qin, Tao and Lu, Jianfeng and Liu, Tie-Yan},
  journal={Advances in neural information processing systems},
  volume={33},
  pages={16857--16867},
  year={2020}
}

@article{liu2019roberta,
  title={Roberta: A robustly optimized bert pretraining approach},
  author={Liu, Yinhan and Ott, Myle and Goyal, Naman and Du, Jingfei and Joshi, Mandar and Chen, Danqi and Levy, Omer and Lewis, Mike and Zettlemoyer, Luke and Stoyanov, Veselin},
  journal={arXiv preprint arXiv:1907.11692},
  year={2019}
}

@inproceedings{nguyen-etal-2020-dynamic,
    title = "Dynamic Semantic Matching and Aggregation Network for Few-shot Intent Detection",
    author = "Nguyen, Hoang  and
      Zhang, Chenwei  and
      Xia, Congying  and
      Yu, Philip",
    editor = "Cohn, Trevor  and
      He, Yulan  and
      Liu, Yang",
    booktitle = "Findings of the Association for Computational Linguistics: EMNLP 2020",
    month = nov,
    year = "2020",
    address = "Online",
    publisher = "Association for Computational Linguistics",
    url = "https://aclanthology.org/2020.findings-emnlp.108",
    doi = "10.18653/v1/2020.findings-emnlp.108",
    pages = "1209--1218",
}

@inproceedings{qi-etal-2021-benchmarking,
    title = "Benchmarking Commercial Intent Detection Services with Practice-Driven Evaluations",
    author = "Qi, Haode  and
      Pan, Lin  and
      Sood, Atin  and
      Shah, Abhishek  and
      Kunc, Ladislav  and
      Yu, Mo  and
      Potdar, Saloni",
    editor = "Kim, Young-bum  and
      Li, Yunyao  and
      Rambow, Owen",
    booktitle = "Proceedings of the 2021 Conference of the North American Chapter of the Association for Computational Linguistics: Human Language Technologies: Industry Papers",
    month = jun,
    year = "2021",
    address = "Online",
    publisher = "Association for Computational Linguistics",
    url = "https://aclanthology.org/2021.naacl-industry.38",
    doi = "10.18653/v1/2021.naacl-industry.38",
    pages = "304--310",
}

@inproceedings{zhou-etal-2023-towards-open,
    title = "Towards Open Environment Intent Prediction",
    author = "Zhou, Yunhua  and
      Hong, Jiawei  and
      Qiu, Xipeng",
    editor = "Rogers, Anna  and
      Boyd-Graber, Jordan  and
      Okazaki, Naoaki",
    booktitle = "Findings of the Association for Computational Linguistics: ACL 2023",
    month = jul,
    year = "2023",
    address = "Toronto, Canada",
    publisher = "Association for Computational Linguistics",
    url = "https://aclanthology.org/2023.findings-acl.140",
    doi = "10.18653/v1/2023.findings-acl.140",
    pages = "2226--2240",
}

@inproceedings{cui2017superagent,
  title={Superagent: A customer service chatbot for e-commerce websites},
  author={Cui, Lei and Huang, Shaohan and Wei, Furu and Tan, Chuanqi and Duan, Chaoqun and Zhou, Ming},
  booktitle={Proceedings of ACL 2017, system demonstrations},
  pages={97--102},
  year={2017}
}

@Misc{peft,
  title =        {PEFT: State-of-the-art Parameter-Efficient Fine-Tuning methods},
  author =       {Sourab Mangrulkar and Sylvain Gugger and Lysandre Debut and Younes Belkada and Sayak Paul and Benjamin Bossan},
  howpublished = {\url{https://github.com/huggingface/peft}},
  year =         {2022}
}

@article{hu2022lora,
  title={Lora: Low-rank adaptation of large language models.},
  author={Hu, Edward J and Shen, Yelong and Wallis, Phillip and Allen-Zhu, Zeyuan and Li, Yuanzhi and Wang, Shean and Wang, Lu and Chen, Weizhu and others},
  journal={ICLR},
  volume={1},
  number={2},
  pages={3},
  year={2022}
}

@misc{guu2020realmretrievalaugmentedlanguagemodel,
      title={REALM: Retrieval-Augmented Language Model Pre-Training}, 
      author={Kelvin Guu and Kenton Lee and Zora Tung and Panupong Pasupat and Ming-Wei Chang},
      year={2020},
      eprint={2002.08909},
      archivePrefix={arXiv},
      primaryClass={cs.CL},
      url={https://arxiv.org/abs/2002.08909}, 
}

@misc{karpukhin2020densepassageretrievalopendomain,
      title={Dense Passage Retrieval for Open-Domain Question Answering}, 
      author={Vladimir Karpukhin and Barlas Oğuz and Sewon Min and Patrick Lewis and Ledell Wu and Sergey Edunov and Danqi Chen and Wen-tau Yih},
      year={2020},
      eprint={2004.04906},
      archivePrefix={arXiv},
      primaryClass={cs.CL},
      url={https://arxiv.org/abs/2004.04906}, 
}

@misc{ram2023incontextretrievalaugmentedlanguagemodels,
      title={In-Context Retrieval-Augmented Language Models}, 
      author={Ori Ram and Yoav Levine and Itay Dalmedigos and Dor Muhlgay and Amnon Shashua and Kevin Leyton-Brown and Yoav Shoham},
      year={2023},
      eprint={2302.00083},
      archivePrefix={arXiv},
      primaryClass={cs.CL},
      url={https://arxiv.org/abs/2302.00083}, 
}

@misc{shuster2022blenderbot3deployedconversational,
      title={BlenderBot 3: a deployed conversational agent that continually learns to responsibly engage}, 
      author={Kurt Shuster and Jing Xu and Mojtaba Komeili and Da Ju and Eric Michael Smith and Stephen Roller and Megan Ung and Moya Chen and Kushal Arora and Joshua Lane and Morteza Behrooz and William Ngan and Spencer Poff and Naman Goyal and Arthur Szlam and Y-Lan Boureau and Melanie Kambadur and Jason Weston},
      year={2022},
      eprint={2208.03188},
      archivePrefix={arXiv},
      primaryClass={cs.CL},
      url={https://arxiv.org/abs/2208.03188}, 
}

@inproceedings{shuster-etal-2022-language,
    title = "Language Models that Seek for Knowledge: Modular Search {\&} Generation for Dialogue and Prompt Completion",
    author = "Shuster, Kurt  and
      Komeili, Mojtaba  and
      Adolphs, Leonard  and
      Roller, Stephen  and
      Szlam, Arthur  and
      Weston, Jason",
    editor = "Goldberg, Yoav  and
      Kozareva, Zornitsa  and
      Zhang, Yue",
    booktitle = "Findings of the Association for Computational Linguistics: EMNLP 2022",
    month = dec,
    year = "2022",
    address = "Abu Dhabi, United Arab Emirates",
    publisher = "Association for Computational Linguistics",
    url = "https://aclanthology.org/2022.findings-emnlp.27/",
    doi = "10.18653/v1/2022.findings-emnlp.27",
    pages = "373--393"
}

@inproceedings{zhuang-etal-2021-weakly,
    title = "Weakly Supervised Extractive Summarization with Attention",
    author = "Zhuang, Yingying  and
      Lu, Yichao  and
      Wang, Simi",
    editor = "Li, Haizhou  and
      Levow, Gina-Anne  and
      Yu, Zhou  and
      Gupta, Chitralekha  and
      Sisman, Berrak  and
      Cai, Siqi  and
      Vandyke, David  and
      Dethlefs, Nina  and
      Wu, Yan  and
      Li, Junyi Jessy",
    booktitle = "Proceedings of the 22nd Annual Meeting of the Special Interest Group on Discourse and Dialogue",
    month = jul,
    year = "2021",
    address = "Singapore and Online",
    publisher = "Association for Computational Linguistics",
    url = "https://aclanthology.org/2021.sigdial-1.54/",
    doi = "10.18653/v1/2021.sigdial-1.54",
    pages = "520--529",
}

@inproceedings{10.1145/3543873.3587680,
author = {Zhuang, Yingying and Song, Jiecheng and Sadagopan, Narayanan and Beniwal, Anurag},
title = {Self-supervised Pre-training and Semi-supervised Learning for Extractive Dialog Summarization},
year = {2023},
isbn = {9781450394192},
publisher = {Association for Computing Machinery},
address = {New York, NY, USA},
url = {https://doi.org/10.1145/3543873.3587680},
doi = {10.1145/3543873.3587680},
booktitle = {Companion Proceedings of the ACM Web Conference 2023},
pages = {1069–1076},
numpages = {8},
location = {Austin, TX, USA},
series = {WWW '23 Companion}
}

@article{schuurmans2019intent,
  title={Intent classification for dialogue utterances},
  author={Schuurmans, Jetze and Frasincar, Flavius},
  journal={IEEE Intelligent Systems},
  volume={35},
  number={1},
  pages={82--88},
  year={2019},
  publisher={IEEE}
}

@inproceedings{ahmadvand2020jointmap,
  title={Jointmap: joint query intent understanding for modeling intent hierarchies in e-commerce search},
  author={Ahmadvand, Ali and Kallumadi, Surya and Javed, Faizan and Agichtein, Eugene},
  booktitle={Proceedings of the 43rd International ACM SIGIR Conference on Research and Development in Information Retrieval},
  pages={1509--1512},
  year={2020}
}

@article{liu2024lara,
  title={LARA: Linguistic-Adaptive Retrieval-Augmentation for Multi-Turn Intent Classification},
  author={Liu, Junhua and Tan, Yong Keat and Fu, Bin and Lim, Kwan Hui},
  journal={arXiv preprint arXiv:2403.16504},
  year={2024}
}

@inproceedings{wang2021mell,
  title={Mell: Large-scale extensible user intent classification for dialogue systems with meta lifelong learning},
  author={Wang, Chengyu and Pan, Haojie and Liu, Yuan and Chen, Kehan and Qiu, Minghui and Zhou, Wei and Huang, Jun and Chen, Haiqing and Lin, Wei and Cai, Deng},
  booktitle={Proceedings of the 27th ACM SIGKDD conference on knowledge discovery \& data mining},
  pages={3649--3659},
  year={2021}
}

@article{yu2021few,
  title={Few-shot intent classification and slot filling with retrieved examples},
  author={Yu, Dian and He, Luheng and Zhang, Yuan and Du, Xinya and Pasupat, Panupong and Li, Qi},
  journal={arXiv preprint arXiv:2104.05763},
  year={2021}
}

@article{pi2024universal,
  title={Universal model in online customer service},
  author={Pi, Shu-Ting and Hsieh, Cheng-Ping and Liu, Qun and Zhu, Yuying},
  journal={Companion Proceedings of the ACM Web Conference 2023},
  year={2023}
}

@article{pi2024topology,
  title={Uncovering customer issues through topological natural language analysis},
  author={Pi, Shu-Ting and Srinivasan, Sidarth and Zhu, Yuying and Yang, Michael and Liu, Qun},
  journal={arXiv:2403.00804},
  year={2024}
}

@article{gupta2024dard,
  title={DARD: A multi-agent approach for task-oriented dialog systems},
  author={Gupta, Aman and Ravichandran, Anirudh and Zhang, Ziji and Shah, Swair and Beniwal, Anurag and Sadagopan, Narayanan},
  journal={arXiv preprint arXiv:2411.00427},
  year={2024}
}
\clearpage
\end{document}